\documentclass[10pt,twocolumn,letterpaper]{article}

\usepackage{wacv}
\usepackage{times}
\usepackage{epsfig}
\usepackage{graphicx}
\usepackage{amsmath}

\usepackage{subcaption}
\usepackage{textcomp }
\usepackage{algorithm}
\usepackage[algo2e,linesnumbered,vlined,ruled]{algorithm2e}
\usepackage{dingbat}
\usepackage{setspace}

\wacvfinalcopy 

\def\wacvPaperID{****} 

\ifwacvfinal\fi
\begin{document}

\title{Mask R-CNN with Pyramid Attention Network for Scene Text Detection}

\author{Zhida Huang\textsuperscript{1,3,$^\dag$,\thanks{Equal contribution. $^\dag$This work was done when Zhida Huang and Zhuoyao Zhong were interns in Speech Group, Microsoft Research Asia, Beijing, China.}}, Zhuoyao Zhong\textsuperscript{2,3,$^\dag$,\footnotemark[1]}, Lei Sun\textsuperscript{3}, Qiang Huo\textsuperscript{3} \\
	\textsuperscript{1}School of Software \& Microelectronics, Peking University \\
	\textsuperscript{2}School of Electronic and Information Engineering, South China University of Technology \\
	\textsuperscript{3}Microsoft Research Asia \\
	\tt\small {hzhida@pku.edu.cn, zhuoyao.zhong@gmail.com, \{lsun, qianghuo\}@microsoft.com}}

\maketitle
\ifwacvfinal\thispagestyle{empty}\fi

\begin{abstract}
   In this paper, we present a new Mask R-CNN based text detection approach which can robustly detect multi-oriented and curved text from natural scene images in a unified manner. To enhance the feature representation ability of Mask R-CNN for text detection tasks, we propose to use the Pyramid Attention Network (PAN) as a new backbone network of Mask R-CNN. Experiments demonstrate that PAN can suppress false alarms caused by text-like backgrounds more effectively. Our proposed approach has achieved superior performance on both multi-oriented (ICDAR-2015, ICDAR-2017 MLT) and curved (SCUT-CTW1500) text detection benchmark tasks by only using single-scale and single-model testing.
\end{abstract}

\section{Introduction}
Scene text detection has drawn increasing attentions from the computer vision community \cite{he2017deep,liao2017textboxes, liu2017deep, Shi2017Detecting, Zhou2017EAST} since it has a wide range of applications in document analysis, robot navigation, OCR translation, image retrieval and augmented reality. However, because of diverse text variabilities in colors, fonts, orientations, languages and scales, extremely complex and text-like backgrounds, as well as some distortions and artifacts caused by image capturing like non-uniform illumination, low contrast, low resolution and occlusion, text detection in natural scene images is still an unsolved problem.

Nowadays, with the astonishing development of deep learning, great progress has been made in this field. Lots of state-of-the-art convolutional neural network (CNN) based object detection and segmentation frameworks, such as Faster R-CNN \cite{ren2015faster}, SSD \cite{liu2016ssd} and FCN \cite{Long2015FCN}, have been borrowed to solve the text detection problem and substantially outperform traditional MSER \cite{matas2004robust} or SWT \cite{epshtein2010detecting} based bottom-up text detection approaches. For example, some approaches \cite{Zhang2016CVPR,Yao2016SceneTD} formulate text detection as a semantic segmentation problem and employ an FCN to make a pixel-level text/non-text prediction, based on which a text saliency map can be generated. As only coarse text-blocks can be detected from the saliency map, complex post-processing steps are needed to extract accurate bounding boxes of text-lines. Unlike FCN-based methods, another category of methods treats text as a specific object and leverages effective object detection frameworks like R-CNN \cite{girshick2014rich}, Faster R-CNN \cite{ren2015faster}, SSD \cite{liu2016ssd}, YOLO \cite{redmon2016you} and DenseBox \cite{huang2015densebox} to detect words or text-lines from images directly. Although these approaches are composed of simpler pipelines, they still struggle with curved text detection. To solve this problem, some recent approaches like PixelLink \cite{Deng2018PixelLink}, FTSN \cite{dai2017fused}, and IncepText \cite{yang2018inceptext}, propose to formulate text detection as an instance segmentation problem so that both straight text and curved text can be detected in a unified manner. Specifically, PixelLink proposes to detect text by linking pixels within the same text instances together, while FTSN and IncepText borrow the FCIS framework \cite{li2016fully} to solve the text detection problem. Although promising results have been achieved, the used instance segmentation approaches have now been surpassed by the latest state-of-the-art Mask R-CNN approach on general instance segmentation tasks \cite{he2017mask}. Therefore, it is straightforward to use Mask R-CNN to further improve the text detection performance.

In this paper, we present an effective Mask R-CNN based text detection approach which can detect multi-oriented and curved text from natural scene images in a unified manner. To enhance the feature representation ability of Mask R-CNN, we propose to use the Pyramid Attention Network (PAN) \cite{Li2018Pyramid} as a new backbone network of Mask R-CNN. Experiments demonstrate that PAN can suppress false alarms caused by text-like backgrounds more effectively. Our proposed approach has achieved superior performance on both multi-oriented (ICDAR-2015 \cite{karatzas2015icdar}, ICDAR-2017 MLT \cite{Nayef2018ICDAR2017}) and curved (SCUT-CTW1500 \cite{Yuliang2017Detecting}) text detection benchmark tasks.

\section{Related work}
In this section, we focus on reviewing recently proposed CNN based text detection approaches and recent developments in instance segmentation tasks.

\subsection{Text Detection}
State-of-the-art CNN based object detection and segmentation frameworks have been widely used to solve the text detection problem recently. Some of these methods \cite{Zhang2016CVPR,Yao2016SceneTD} borrow the idea of semantic segmentation and employ an FCN to make a pixel-level text/non-text prediction, which produces a text saliency map for text detection. However, only coarse text-blocks can be detected from this saliency map, so complex post-processing steps are needed to extract accurate bounding boxes of text-lines. Another category of methods \cite{Jaderberg2016RCNN,Gupta_2016_CVPR,Zhong2017DeepText,liao2017textboxes,ma2018arbitrary,liu2017deep,Zhou2017EAST,he2017deep} treats text as a specific object and leverages state-of-the-art object detection frameworks to detect word or text-lines from images directly. Jaderberg et al. \cite{Jaderberg2016RCNN} adapted R-CNN for text detection, while its performance was limited by the traditional region proposal generation methods. Gupta et al. \cite{Gupta_2016_CVPR} borrowed the YOLO framework and employed a fully-convolutional regression network to perform text detection and bounding box regression at all locations and multiple scales of an image. Zhong et al. \cite{Zhong2017DeepText} and Liao et al. \cite{liao2017textboxes} employed the Faster R-CNN and SSD frameworks to solve the word-level horizontal text detection problem, respectively. In order to extend Faster R-CNN and SSD to multi-oriented text detection, Ma et al. \cite{ma2018arbitrary} and Liu et al. \cite{liu2017deep} proposed quadrilateral anchors to hunt for inclined text proposals which could better fit the multi-oriented text instances. To overcome the inefficiency of anchor mechanism \cite{he2017deep}, Zhou et al. \cite{Zhou2017EAST} and He et al. \cite{he2017deep} borrowed the idea of DenseBox and used a one-stage FCN to output pixel-wise textness scores as well as the quadrilateral bounding boxes through all locations and scales of an image. Although these approaches are composed of simpler pipelines, they still struggle with curved text detection. Recently, instead of detecting the whole words or text-lines directly, Tian et al. \cite{tian2016detecting} and Shi et al. \cite{Shi2017Detecting} adopted object detection methods to detect text segments firstly, then grouped these text segments into words or lines with some simple text-line grouping algorithms or the learned linkage information, respectively. Intuitively, these methods can be applied for curved text detection, but they make the total text detection pipeline more sophisticated.  Moreover, the segment grouping problem itself is a nontrivial problem, especially when the layout is complex, e.g., text with large character spacing, which will affect the text detection performance too. To overcome the above problems, some recent approaches propose to formulate text detection as an instance segmentation problem so that both straight text and curved text can be detected in a unified manner. Deng et al. \cite{Deng2018PixelLink} proposed to detect text by linking pixels within the same text instances together. Dai et al. \cite{dai2017fused} and Yang et al. \cite{yang2018inceptext} adopted the FCIS framework \cite{li2016fully} to solve the text detection problem. In this paper, we borrowed Mask R-CNN, which is the latest state-of-the-art instance segmentation approach, to further enhance the text detection performance.

\subsection{Instance Segmentation}
Instance segmentation is a challenging task because it requires the correct detection of all objects in an image while also precisely segmenting each instance. Dai et al. \cite{dai2016instance} proposed a complex multiple-stage cascade that predicts segment proposals from bounding-box proposals, followed by classification. Later, Li et al. \cite{li2016fully} combined the segment proposal system in \cite{Dai2016Instance-Sensitive} and R-FCN \cite{dai2016r} for fully convolutional instance segmentation (FCIS). Although fast, FCIS exhibits systematic errors on overlapping instances and creates spurious edges \cite{he2017mask}. More recently, Mask R-CNN \cite{he2017mask} extended Faster R-CNN \cite{ren2015faster} by adding a branch for predicting an object mask in parallel with the existing branch for bounding box recognition. It introduced RoIAlign \cite{he2017mask} to replace RoIPool \cite{girshick2015fast} to fix the pixel misalignment and used ResneXt \cite{xie2017aggregated} as the base network. Moreover, it took advantage of Feature Pyramid Network (FPN \cite{lin2017feature}) to strengthen feature representation ability and partially eased the problem of small object detection. In this paper, to further enhance the feature representation ability of Mask R-CNN, we propose to incorporate the Pyramid Attention Network (PAN) \cite{Li2018Pyramid} into the Mask R-CNN framework. Experiments demonstrate that PAN can suppress false alarms caused by text-like backgrounds more effectively.

\begin{figure*}
	\begin{center}
		\includegraphics[width=1.0\linewidth, height=7cm]{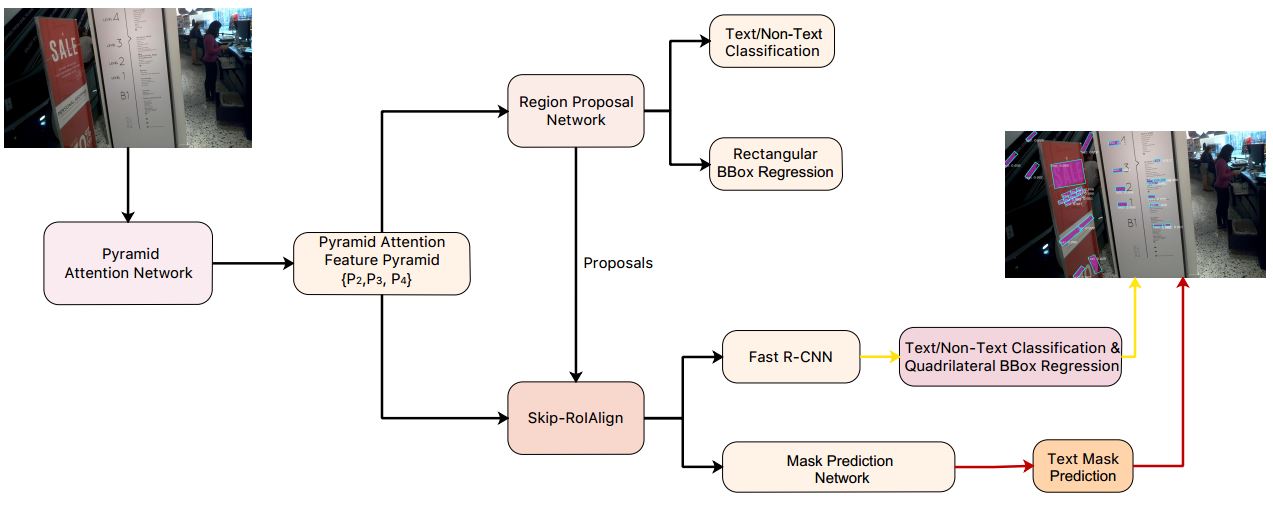}
	\end{center}
	\caption{Architecture of our Mask R-CNN based text detector, which consists of a PAN backbone network, a region proposal network, a Fast R-CNN detector and a mask prediction network.}
	\label{fig:short}
\end{figure*}

\section{Our Method}
Our Mask R-CNN based text detection network is composed of four modules: 1) A PAN backbone network that is responsible for computing a multi-scale convolutional feature pyramid over a full image; 2) A region proposal network (RPN) that generates rectangular text proposals; 3) A Fast R-CNN detector that classifies extracted proposals and outputs the corresponding quadrilateral bounding boxes; 4) A mask prediction network that predicts text masks for input proposals. A schematic view of our text detection network is depicted in Fig. 1 and details are described in the following subsections.

\subsection{Pyramid Attention Network}

Recently, Li et al. \cite{Li2018Pyramid} proposed a Pyramid Attention Network (PAN) that combines the attention mechanism and spatial pyramid to extract precise dense features for semantic segmentation tasks. It mainly consists of two modules, i.e., a Feature Pyramid Attention (FPA) module and a Global Attention Up-sample (GAU) module. The FPA module performs spatial pyramid attention on high-level features and combines global pooling to learn better high-level feature representations. The GAU module is attached on each decoder layer to provide global context as a guidance of low-level features to select category localization details. Owing to these tactful designs, PAN achieves state-of-the-art segmentation performance on the VOC2012 and Cityscapes benchmark tasks. Inspired by this, we propose to use PAN as a new backbone network to improve the feature representation learning for our Mask R-CNN based text detection model.

We build PAN on top of ResNet50 \cite{he2016deep} and ResNeXt50 \cite{xie2017aggregated}. The implementations of PAN generally follow \cite{Li2018Pyramid} with just some modest modifications. As shown in Fig. 2, our FPA module takes the output features of the Res-4 layers in ResNet50 or ResNeXt50 as input, on which it performs 3$\times$3 dilated convolution with sampling rates 3, 6, 12 respectively to better extract context information. These three feature maps are then concatenated and dimension reduced by a 1$\times$1 convolution layer. After that, FPA performs a 1$\times$1 convolution on the input Res-4 features further, whose output is multiplied with the above context features in a pixel-wise manner. The extracted features are added with the output features of the global pooling branch to get the final pyramid attention features. The GAU module, as is shown in Fig. 3, performs 3$\times$3 convolution on the low-level features to reduce channels of feature maps from CNNs. The global context generated from high-level features is through a 1$\times$1 convolution with instance normalization \cite{Ulyanov2016Instance} and ReLU nonlinearity, then multiplied by the low-level features.  Finally, the high-level features after up-sampling are added with the weighted low-level features to generate the GAU features. With the above FPA and GAU modules, we construct a powerful feature pyramid with three levels, i.e., $P_{2}$, $P_{3}$ and $P_{4}$, whose strides are 4, 8 and 16, respectively. The overall PAN architecture is depicted in Fig. 4. We refer readers to \cite{Li2018Pyramid} for further details.

\begin{figure}[t]
	\begin{center}
		\includegraphics[width=0.9\linewidth]{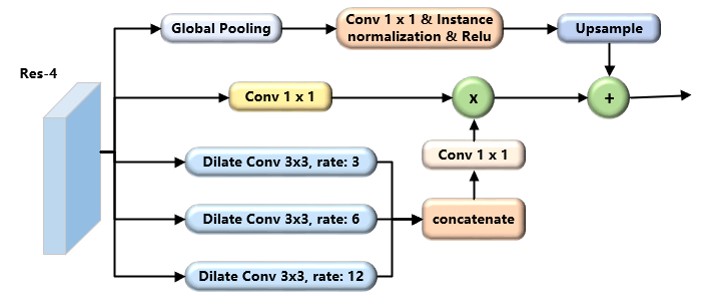}
	\end{center}
	\caption{FPA of our PAN.}
	\label{fig:long}
	\label{fig:onecol}
\end{figure}

\subsection{Region Proposal Network}
Three RPNs are attached to $P_{2}$, $P_{3}$ and $P_{4}$ respectively, each of which slides a small network densely on the corresponding pyramid level to perform text/non-text classification and bounding box regression. The small network is implemented as a 3$\times$3 convolutional layer followed by two sibling 1$\times$1 convolutional layers, which are used for predicting textness score and rectangular bounding box locations respectively. As the size and aspect ratio variabilities of scene text instances are wider than general objects, we design a complicated set of anchors following \cite{liao2017textboxes}. Specifically, we design 6 anchors at each sliding position on each pyramid level in \{$P_{2}$, $P_{3}$, $P_{4}$\} by using 6 aspect ratios \{0.2, 0.5, 1.0, 2.0, 4.0, 8.0\} and one scale in \{32, 64, 128\}. The detection results of all three RPNs are aggregated together to construct a proposal set $\{D\}$. Then, we use the standard non-maximum suppression (NMS) algorithm with an IoU threshold of 0.7 to remove redundant proposals in $\{D\}$, and select the top-$N$ scoring proposals for the succeeding Fast R-CNN and mask prediction network. $N$ is set to 2000 in both the training and testing stages.

\begin{figure}
	\begin{center}
		\includegraphics[width=0.9\linewidth]{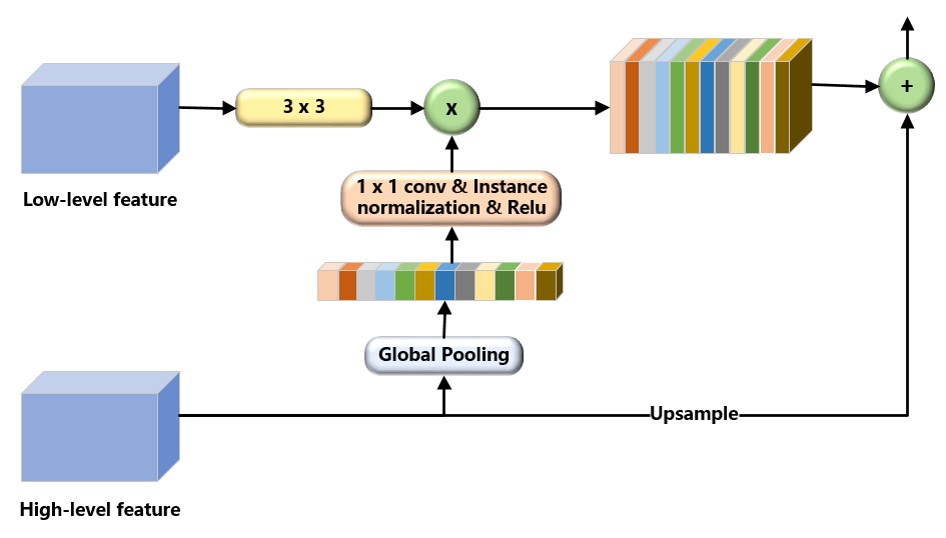}
	\end{center}
	\caption{GAU of our PAN.}
	\label{fig:long}
	\label{fig:onecol}
\end{figure}

\begin{figure}[t]
	\begin{center}
		\includegraphics[width=0.8\linewidth]{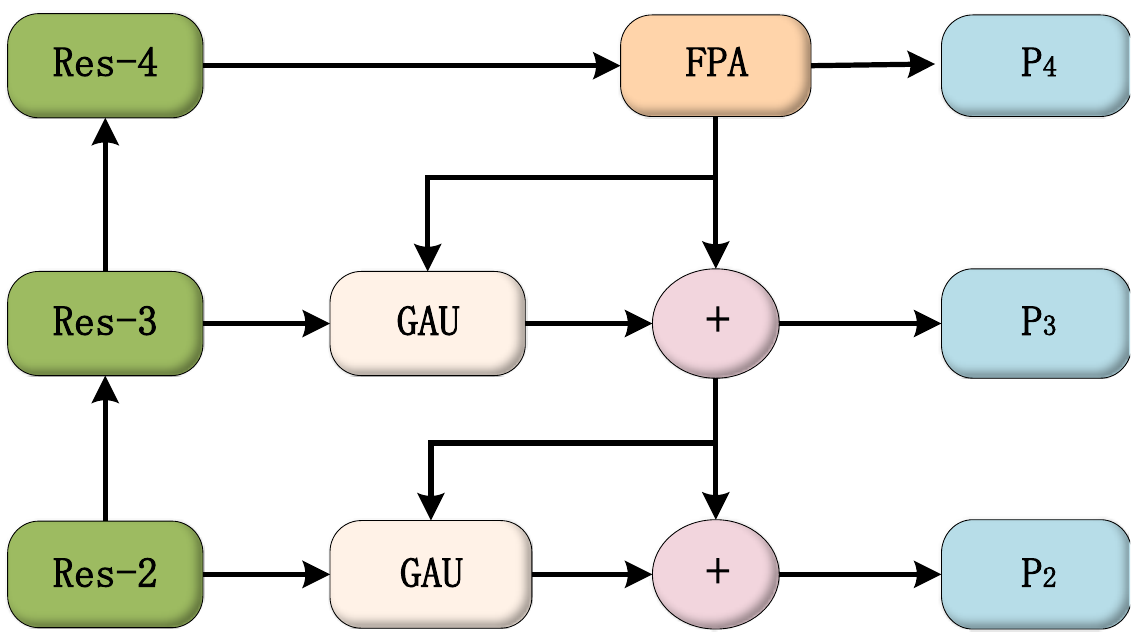}
	\end{center}
	\caption{Architecture of our PAN.}
	\label{fig:long}
	\label{fig:onecol}
\end{figure}

\subsection{Fast R-CNN \& Mask Prediction Network}

After the region proposal generation step, extracting effective features for each proposal is critical to the performance of the following Fast R-CNN and mask prediction network. In the original Faster R-CNN \cite{ren2015faster}, the features of all proposals are extracted from the last convolution layer of the backbone network, which would lead to insufficient features for small proposals. In the recent FPN \cite{lin2017feature}, the features of proposals are extracted from different pyramid levels according to their sizes, i.e., the features of small proposals are extracted from low-level pyramid levels, while large proposals from high-level pyramid levels. Although more effective, there still exists room for further improvement \cite{PANet}. As the features from the $P_{2}$ and $P_{3}$ levels have higher resolution and contain more detailed information, which are complementary to more abstract but low-resolution features from the $P_{4}$ level, it is straightforward to combine these three pyramid levels together to improve the feature representation ability. To achieve this, we borrow the idea of ION \cite{ION} and propose a Skip-RoIAlign method to fuse the $P_{2}$, $P_{3}$ and $P_{4}$ levels. Concretely, for each proposal, we apply ROIAlign over $P_{2}$, $P_{3}$ and $P_{4}$ pyramid levels respectively and extract three feature descriptors with a fixed spatial size of 7$\times$7, which are concatenated and dimension reduced with a 1$\times$1 convolutional layer to obtain the final ROI features. These ROI features are then fed into the network head for text/non-text classification, quadrilateral bounding box regression and mask prediction. Details of the network head are depicted in Fig. 5. The head includes the 5-th stage of ResNet50 or ResNeXt50, which is shared by the Fast R-CNN and mask prediction network.

\begin{figure}[t]
	\begin{center}
		\includegraphics[width=1.0\linewidth]{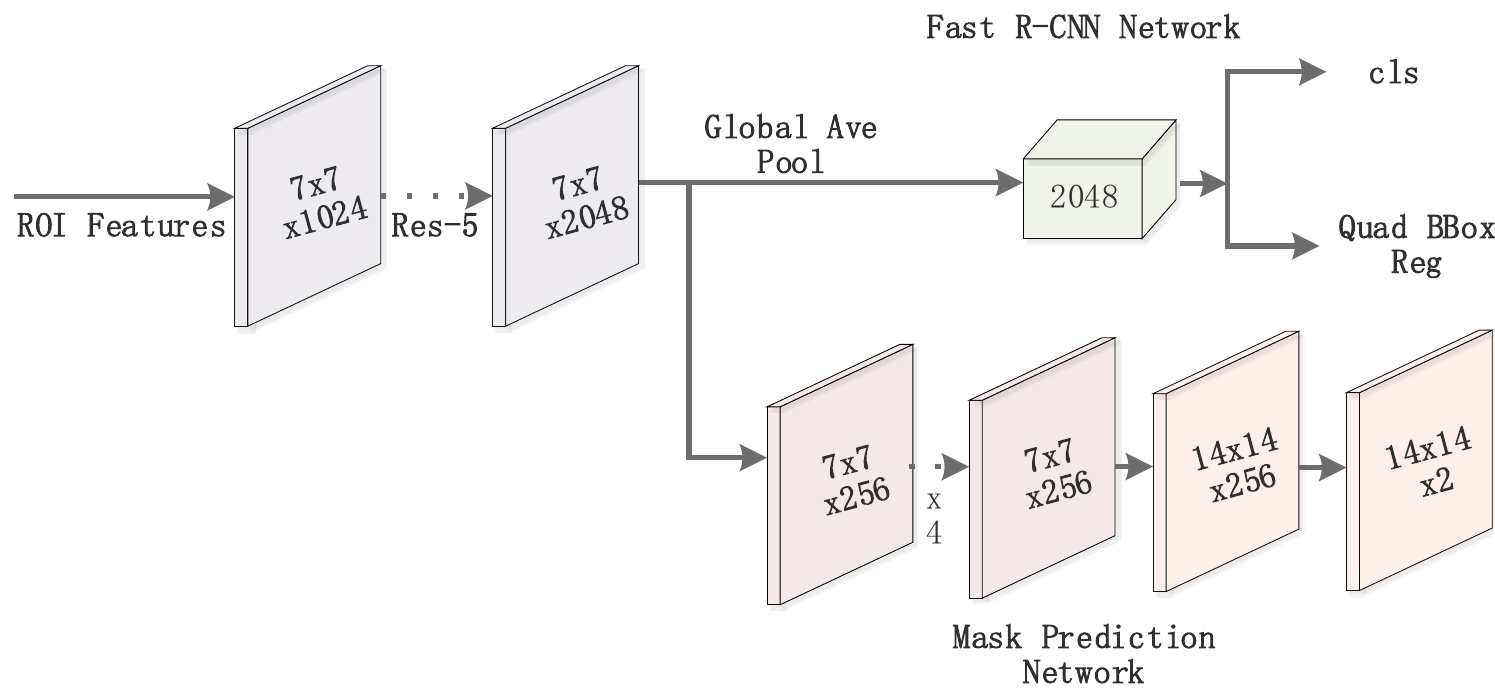}
	\end{center}
	\caption{Architecture of Fast R-CNN and mask prediction network. The ROI features are first fed into the the 5-th stage of ResNet50 or ResneXt50, whose output features are shared by both Fast R-CNN and mask prediction network. Then, for Fast R-CNN, the output features are globally average pooled before the final text/non-text classification and quadrilateral bounding box regression layers, while for the mask prediction network, the output features are followed by four consecutive 3$\times$3 convolutional layers and then up-sampled before the final mask prediction layers. Numbers denote spatial resolution and channels.}
	\label{fig:long}
	\label{fig:onecol}
\end{figure}

\subsection{Training}
\subsubsection{Loss Functions}
\noindent{\textbf{Multi-task loss for RPN. }}There are two sibling output layers for each individual RPN, i.e., a text/non-text classification layer and a rectangular bounding box regression layer. The multi-task loss function can be denoted as follows:
\begin{equation}
L_{RPN_{p_i}}=L^{R}_{cls}(c,c^*)+\lambda_{loc}L^{R}_{loc}(r,r^*),
\end{equation}
where $c$ and $c^*$ are predicted and ground-truth labels respectively, $L^{R}_{cls}(c,c^*)$ is a softmax loss for classification tasks; $r$ and $r^*$ represent the predicted and ground-truth 4-dimensional parameterized regression targets as stated in \cite{ren2015faster}, $L^{R}_{loc}(r,r^*)$ is a smooth-$L_1$ loss \cite{girshick2015fast} for regression tasks. $\lambda_{loc}$ is a loss-balancing parameter, and we set $\lambda_{loc}=3$.

The total loss of RPN $L_{RPN}$ is the sum of the losses of the three RPNs.

\noindent{\textbf{Multi-task loss for Fast R-CNN. }}Fast R-CNN also has two sibling output layers: 1) A text/non-text classification layer, which is the same as the above-mentioned RPN; 2) A quadrilateral bounding box regression layer. The multi-task loss function for Fast R-CNN is defined as follows:
\begin{equation}
L_{FRCN}=L^{F}_{cls}(c,c^*)+\lambda_{loc}L^{F}_{loc}(t,t^*),
\end{equation}
where $t=\{$$(\bigtriangleup_{x_i},\bigtriangleup_{y_i})|i\in\{1,2,3,4\}\}$ and $t^*=\{$$(\bigtriangleup^*_{x_i},\bigtriangleup^*_{y_i})|i\in\{1,2,3,4\}\}$ represent the predicted and ground-truth 8-dimensional parameterized coordinate offsets. Let $\{(x^{g}_i,y^{g}_i)|i\in\{1,2,3,4\}\}$ denote the four vertices of $G$ and $(x^{p}_1,y^{p}_1,x^{p}_2,y^{p}_2,P_w, P_h)$ be the top-left and bottom-right coordinates, width and height of an input proposal $P$. The parameterizations of $t^*$ are denoted as:
\begin{align*}
&\triangle^*_{x_1} = (x_1^g - x_1^p) / P_w ,\quad \triangle^*_{y_1} = (y_1^g - y_1^p) / P_h ,\\
~& \triangle^*_{x_2} = (x_2^g - x_2^p) / P_w , \quad \triangle^*_{y_2} = (y_2^g - y_1^p) / P_h ,  \\
~& \triangle^*_{x_3} = (x_3^g - x_2^p) / P_w , \quad \triangle^*_{y_3} = (y_3^g - y_2^p) / P_h ,\\
~& \triangle^*_{x_4} = (x_4^g - x_1^p) / P_w , \quad \triangle^*_{y_4} = (y_4^g - y_2^p) / P_h .  \tag{3}
\end{align*}
$L^{F}_{loc}(t,t^*)$ is also a smooth-$L_1$ loss and we set $\lambda_{loc}=1$.

\noindent{\textbf{Loss for mask prediction network. }}Let $m$ and $m^*$ be the predicted and ground-truth mask targets respectively and $L_{mask}(m,m^*)$ be a standard binary cross-entropy loss for mask prediction tasks. Based on these definitions, the loss function can be defined as follows:
\begin{align*}
L_{MASK}=L_{mask}(m,m^*). \tag{4}
\end{align*}
The overall loss function for training the proposed Mask R-CNN based text detection model can be denoted as:
\begin{align*}
L=L_{RPN}+L_{FRCN}+\lambda_{mask}L_{MASK}, \tag{5}
\end{align*}
where $\lambda_{mask}$ is a loss-balancing parameter for $L_{MASK}$, and we set $\lambda_{mask}=0.03125$.
\subsubsection{Training Details}
In each training iteration of RPN, we sample a mini-batch of 128 positive and 128 negative anchors for each RPN. An anchor is assigned a positive label if it has the highest IoU for a given ground-truth bounding box or has an IoU over 0.7 with any ground-truth bounding box, and a negative label if its IoU overlap is less than 0.3 for all ground-truth bounding boxes. For Fast R-CNN, we sample a mini-batch of 64 positive and 192 negative text proposals in each iteration. A proposal is assigned a positive label if it has an IoU over 0.5 with any ground-truth bounding box, otherwise assigned a negative label. For the sake of efficiency, the IoU overlaps between proposals and ground-truth boxes are calculated using their axis-aligned rectangular bounding boxes. Only the positive text proposals are used for training the mask prediction network. The mask target is the intersection between
a proposal and its associated ground-truth mask.

\section{Experiments}
We evaluate our proposed method on several standard benchmark tasks， including ICDAR-2015 \cite{karatzas2015icdar} and  ICDAR-2017 MLT 2017 \cite{Nayef2018ICDAR2017} for multi-oriented text detection, and SCUT-CTW 1500 \cite{Yuliang2017Detecting} for curved text detection. Text instances are labeled in word-level with quadrilateral bounding boxes in the former two datasets and in text-line level with 14 coordinate points in SCUT-CTW 1500. ICDAR-2017 MLT is built for the multi-lingual scene text detection and script identification challenge in the ICDAR-2017 Robust Reading Competition, which includes 9 languages: Chinese, Japanese, Korean, English, French, Arabic, Italian, German and Indian. It contains 7,200, 1,800 and 9,000 images for training, validation and testing, respectively.  ICDAR-2015 is built for the Incidental Scene Text challenge in the ICDAR-2015 Robust Reading Competition, which contains 1,000 and 500 images for training and testing. SCUT-CTW 1500 is a curved text detection dataset, including 1,000 training images and 500 testing images.

To make our results comparable to others, we use the online official evaluation tools to evaluate the performance of our approach on ICDAR-2017 MLT and ICDAR-2015, and use the evaluation tool provided by the authors of \cite{Yuliang2017Detecting} on SCUT-CTW 1500.

\begin{figure*}
	\begin{center}
		\includegraphics[width=1.0\textwidth, height=6.5cm]{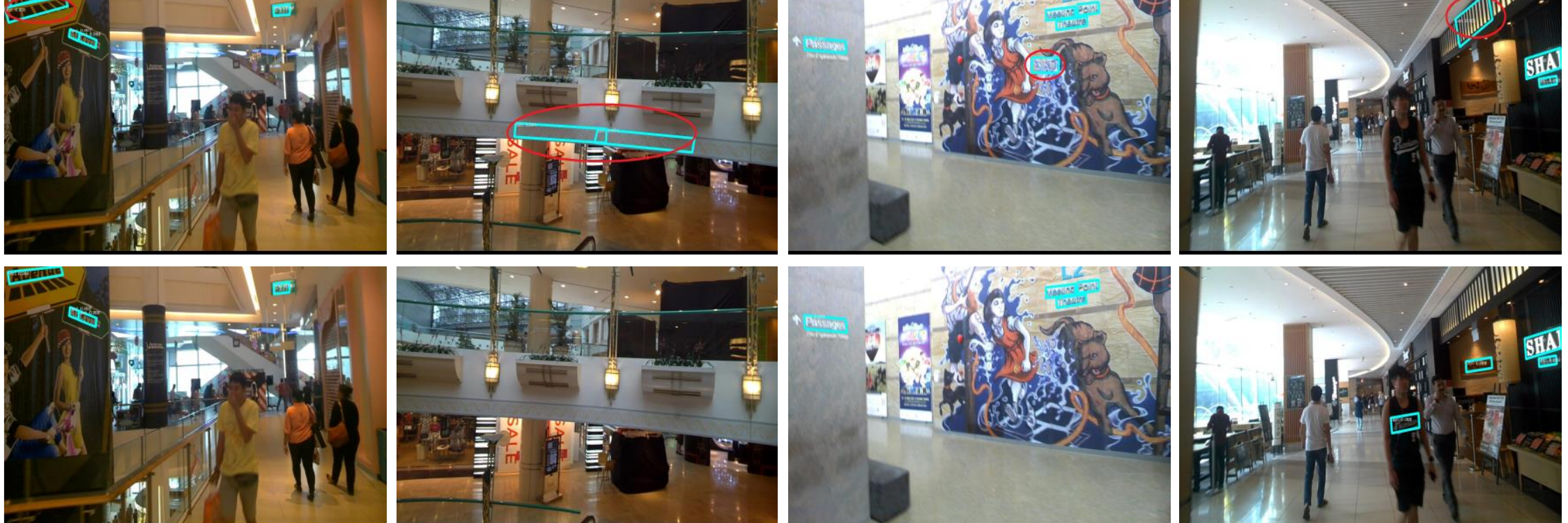}
		\caption{PAN can effectively suppress some false alarms caused by text-like backgrounds. The first row: detection results of Mask R-CNN with FPN. The second row: detection results of Mask R-CNN with PAN.}
	\end{center}
	\label{fig:differnet_accent}       
\end{figure*}

\subsection{Implementation Details}
The weights of ResNet50 or ResNeXt50 related layers in the PAN backbone network are initialized by using the corresponding pre-trained models from the ImageNet classification task \cite{he2016deep,xie2017aggregated}. The weights of the new layers for PAN, RPN, Fast R-CNN and mask prediction network are initialized by using random weights with a Gaussian distribution of mean 0 and standard deviation 0.01. Our Mask R-CNN based text detection model is trained in an end-to-end manner and optimized by the standard SGD algorithm with a momentum of 0.9 and weight decay of 0.0005. For ICDAR-2017 MLT, we use the training and validation data, i.e., a total of 9,000 images for training, while for both ICDAR-2015 and SCUT-CTW 1500, we only use the provided training images for training.

We implement our approach based on MXNet and experiments are conducted on a workstation with 4 Nvidia P100 GPUs. We adopt a multi-scale training strategy. The scale $S$ is defined as the length of the shorter side of an image. In each training iteration, a selected training image is individually rescaled by randomly sampling a scale $S$ from the set \{480, 576, 720, 928, 1088\},
\{480, 576, 688, 720, 928\}, and \{300, 400, 500, 600, 704\} for ICDAR-2017 MLT, ICDAR-2015 and SCUT-CTW 1500, respectively.

In the testing phase, we keep the top-2000 scoring text proposals generated by RPN for the succeeding Fast R-CNN. After the Fast R-CNN step, quadrilateral bounding boxes of detected text instances are predicted and suppressed by the Skewed NMS \cite{ma2018arbitrary} algorithm with an IoU threshold of 0.3. Finally, ROI features in the axis-aligned rectangular bounding box of each remaining text instance are fed into the mask prediction network to get the text mask. For the ICDAR-2017 MLT and ICDAR-2015 datasets, we directly use the quadrilateral bounding boxes predicted by the Fast R-CNN module as the final detection results, while for the curved text detection dataset SCUT-CTW 1500, we use the text masks predicted by the mask prediction network as the final detection results.

\subsection{Component evaluation}
In this section, we conduct a series of ablation experiments to evaluate the effectiveness of the base convolutional network and PAN on ICDAR-2017 MLT, ICDAR-2015 and SCUT-CTW 1500 text detection benchmark datasets. All the experiments are based on single-model and single-scale testing. The scales of testing images are set as 1440, 1024 and 512 for ICDAR-2017 MLT, ICDAR-2015 and SCUT-CTW 1500, respectively.

	\begin{table}
	\begin{center}
		\begin{tabular}{|c|c|c|c|c|c|c|}
			\hline
			Base Network & FPN & PAN & R & P & F \\
			\hline\hline
			ResNet50 & \checkmark &  & \textbf{0.687} & 0.744 & 0.714  \\
			\hline
			ResNet50   &  & \checkmark   & 0.686 & \textbf{0.787} & \textbf{0.733} \\
			\hline\hline
			ResneXt50 & \checkmark &  & 0.686 & 0.795 & 0.737  \\
			\hline
			ResneXt50   &  & \checkmark   & \textbf{0.698} & \textbf{0.800} & \textbf{0.743} \\
			\hline
		\end{tabular}
	\end{center}
	\vspace{-0.4cm}
	\caption{Component evaluation on ICDAR-2017 MLT. R, P and F stand for recall, precision and F-measure respectively.}
\end{table}

	\begin{table}
	\begin{center}
		\begin{tabular}{|c|c|c|c|c|c|c|}
			\hline
			Base Network & FPN & PAN & R & P & F \\
			\hline\hline
			ResNet50 & \checkmark &  & \textbf{0.818} & 0.877 & 0.846  \\
			\hline
			ResNet50   &  & \checkmark   & \textbf{0.818} & \textbf{0.882} & \textbf{0.849} \\
			\hline\hline
			ResneXt50 & \checkmark &  & 0.806 & 0.899 & 0.850  \\
			\hline
			ResneXt50   &  & \checkmark   & \textbf{0.815} & \textbf{0.908} & \textbf{0.859} \\
			\hline
		\end{tabular}
	\end{center}
	\vspace{-0.4cm}
	\caption{Component evaluation on ICDAR-2015. R, P and F stand for recall, precision and F-measure respectively.}
\end{table}

\begin{table}
	\begin{center}
		\begin{tabular}{|c|c|c|c|c|c|c|}
			\hline
			Base Network & FPN & PAN & R & P & F \\
			\hline\hline
			ResneXt50 & \checkmark &  & 0.826 & 0.839 & 0.833  \\
			\hline
			ResneXt50   &  & \checkmark   & \textbf{0.832} & \textbf{0.868} & \textbf{0.850} \\
			\hline
		\end{tabular}
	\end{center}
	\vspace{-0.4cm}
	\caption{Component evaluation on SCUT-CTW 1500. R, P and F stand for recall, precision and F-measure respectively.}
\end{table}

\noindent{\textbf{ResNeXt50 is better than ResNet50.}
	As an important part of a backbone network (e.g., ResNet50-FPN), the base convolutional network (e.g., ResNet50) affects the text detection performance a lot. Here we compare the performance of two different base convolutional networks, i.e., Resnet50 and ResneXt50, on ICDAR-2017 MLT and ICDAR-2015. As shown in Table 1 and Table 2, ResneXt50 can consistently outperform ResNet50. In the following experiments, we will use ResneXt50 as our base convolutional network.
	
	\noindent{\textbf{PAN is more powerful than FPN. }
		We compare PAN with FPN \cite{lin2017feature} on all three datasets. As shown in Tables 1-3, no matter which base network is used, PAN consistently outperforms FPN on all datasets, which can demonstrate the effectiveness of PAN. The major improvement of PAN comes from the higher precision especially when the base network is relatively weaker, e.g., when ResNet50 is used as the base network on ICDAR-2017 MLT, PAN can increase the precision by 4.3\% absolutely (Table 1). Some qualitative comparison examples on ICDAR-2015 are shown Fig. 6, from which we can find that some false alarms caused by text-like backgrounds could be suppressed by PAN.

\begin{figure*}
	\begin{center}
		\includegraphics[width=1.0\textwidth, height=4cm]{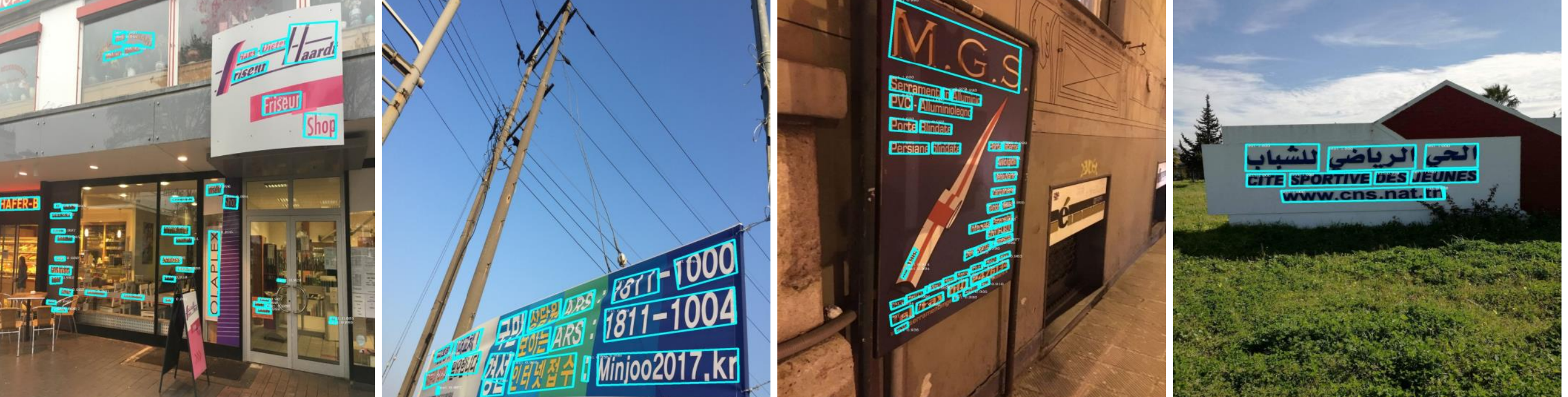}
		\caption{Detection results of our proposed Mask R-CNN based text detector on ICDAR-2017 MLT.}
	\end{center}
	\label{fig:differnet_accent}       
\end{figure*}

\begin{figure*}
	\begin{center}
		\includegraphics[width=1.0\textwidth, height=8.2cm]{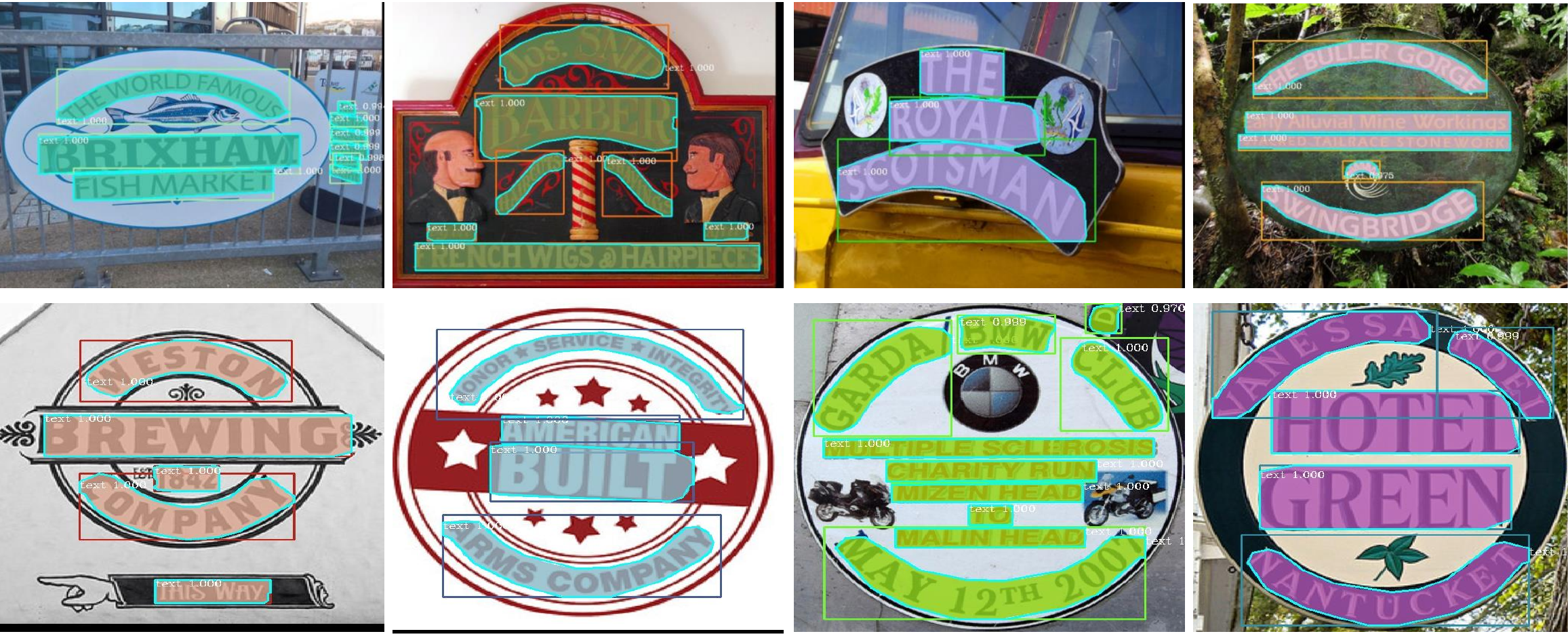}
		\caption{Detection results of our proposed Mask R-CNN based text detector on SCUT-CTW1500.}
	\end{center}
	\label{fig:differnet_accent}       
\end{figure*}

		\subsection{Comparison with Prior Arts}
		We compare the performance of our approach with other most competitive results on the ICDAR-2017 MLT, ICDAR-2015 and SCUT-CTW 1500 text detection benchmark datasets. For fair comparisons, we report all results without using recognition information. As shown in Tables 4-6, our approach achieves the best performance on these three datasets by only using single-scale and single-model testing. Specifically, as shown in Table 4, our approach outperforms the closest method \cite{liu2018fots} significantly by improving the F-measure from 0.707 to 0.743 on the challenging ICDAR-2017 MLT dataset, even though \cite{liu2018fots} applies multi-scale testing to achieve the best possible performance. On the ICDAR-2015 dataset, as shown in Table 5, even though some other approaches have used extra training data, our approach still achieves the best result of 0.815, 0.908 and 0.859 in recall, precision and F-measure respectively. On the SCUT-CTW 1500 dataset, our approach has achieved a new state-of-the-art result, i.e., 0.832, 0.868 and 0.850 in recall, precision and F-measure respectively as shown in Table 6, outperforming other methods by a large margin.  The superior performance achieved by our proposed approach on these three challenging text detection benchmarks can demonstrate the advantage of our approach. Some qualitative detection results are depicted in Figs. 6-8.
		
		\begin{table}
			\begin{center}
				\begin{tabular}{|c|c|c|c|}
					\hline
					Method  & R & P & F \\
					\hline\hline
					Proposed & \textbf{0.698}  & \textbf{0.800} & \textbf{0.743} \\
					\hline
					FOTS MS \cite{liu2018fots} &  0.623 & 0.818  & 0.707 \\
					\hline
					SCUT DLVClab1 \cite{Nayef2018ICDAR2017}  & 0.545 & 0.802   & 0.649 \\
					\hline
					SARI FDU RRPN v1 \cite{ma2018arbitrary} & 0.555 & 71.17   & 0.623 \\
					\hline
					TDN SJTU2017 \cite{Nayef2018ICDAR2017}  & 0.471 & 0.642   & 0.543 \\
					\hline
				\end{tabular}
			\end{center}
			\vspace{-0.4cm}
			\caption{Comparison with prior arts on ICDAR-2017 MLT. R, P and F stand for recall, precision and F-measure respectively. MS indicates using multi-scale testing.}
		\end{table}
		
		\begin{table}
			\begin{center}
				\begin{tabular}{|c|c|c|c|c|}
					\hline
					Method & ExtraData & R & P & F \\
					\hline\hline
					Proposed & \textbf{\Large \texttimes} & \textbf{0.815}  & \textbf{0.908} & \textbf{0.859} \\
					\hline
					IncepText \cite{yang2018inceptext} & \textbf{\Large \texttimes} & 0.806 &  0.905 & 0.853 \\
					\hline
					FTSN \cite{dai2017fused} & \checkmark    & 0.800  & 0.886 & 0.841 \\
					\hline
					R2CNN \cite{jiang2017r2cnn} & \checkmark    & 0.797  & 0.856 & 0.825 \\
					\hline
					DDR \cite{he2017deep} & \checkmark    & 0.800  & 0.820 & 0.810 \\
					\hline
					EAST \cite{Zhou2017EAST} & \textbf{\Large --}   & 0.783  & 0.832 & 0.807 \\
					\hline
					RRPN \cite{ma2018arbitrary} & \checkmark    & 0.732  & 0.822 & 0.774 \\
					\hline
					SegLink \cite{Shi2017Detecting} & \checkmark    & 0.731  & 0.768 & 0.749 \\
					\hline
				\end{tabular}
			\end{center}
			\vspace{-0.4cm}
			\caption{Comparison with prior arts on ICDAR-2015. R, P and F stand for recall, precision and F-measure respectively. }
		\end{table}

		\begin{table}[h]
			\begin{center}
				\begin{tabular}{|c|c|c|c|}
					\hline
					Method  & R & P & F \\
					\hline\hline
					Proposed & \textbf{0.832}  & \textbf{0.868} & \textbf{0.850} \\
					\hline
					CTD+TLOC \cite{Yuliang2017Detecting} & 0.698 &  0.774 & 0.734 \\
					\hline
					DMPNet \cite{liu2017deep}   & 0.560  & 0.699 & 0.622 \\
					\hline
					EAST \cite{Zhou2017EAST} & 0.491  & 0.787 & 0.604 \\
					\hline
					CTPN \cite{tian2016detecting}   & 0.538  & 0.604 & 0.569 \\
					\hline
				\end{tabular}
			\end{center}
			\vspace{-0.4cm}
			\caption{Comparison with prior arts on SCUT-CTW 1500. R, P and F stand for recall, precision and F-measure respectively. }
		\end{table}
		
		\section{Conclusion and Discussion}
		
		A new Mask R-CNN based text detection approach has been proposed in this paper. Thanks to the flexibility of Mask R-CNN, the proposed approach can detect multi-oriented and curved text from natural scene images robustly in a unified manner. Moreover, we demonstrate that using the Pyramid Attention Network (PAN) as a new backbone network of Mask R-CNN enhances the feature representation ability of Mask R-CNN significantly, so that false alarms caused by text-like backgrounds are suppressed more effectively. Our proposed approach has achieved superior performance on both multi-oriented (ICDAR-2015, ICDAR-2017 MLT) and curved (SCUT-CTW1500) text detection benchmark tasks by only using single-scale and single-model testing. However, our approach still has some limitations. First, the running speed of our approach is not fast enough due to the computation intensive PAN backbone network and Mask R-CNN framework. Moreover, our approach struggles with skewed nearby long text-lines owing to the limitation of rectangular proposals generated by RPN. More researches are needed to address these challenging problems.
		
		
		
		{\small
			\bibliographystyle{ieee}
			\bibliography{egpaper_final}
		}

\end{document}